\documentclass[runningheads]{llncs}
\usepackage{graphicx}
\usepackage{float}
\usepackage{eucal}
\usepackage{multirow}
\usepackage{subcaption}
\usepackage{hhline}
\usepackage{colortbl}
\usepackage{tablefootnote}

\makeatletter
\renewcommand\subsubsection{\@startsection{subsubsection}{3}{\z@}%
                       {-18\p@ \@plus -4\p@ \@minus -4\p@}%
                       {0.5em \@plus 0.22em \@minus 0.1em}%
                       {\normalfont\normalsize\bfseries\boldmath}}
\makeatother
 \newcommand\tab[1][1cm]{\hspace*{#1}}
 \usepackage[labelfont=bf]{caption}

\newcommand{\repthanks}[1]{\textsuperscript{\ref{#1}}}
\makeatletter
\patchcmd{\maketitle}
  {\def\thanks}
  {\let\repthanks\repthanksunskip\def\thanks}
  {}{}
\patchcmd{\@maketitle}
  {\def\thanks}
  {\let\repthanks\@gobble\def\thanks}
  {}{}
\newcommand\repthanksunskip[1]{\unskip{}}
\makeatother

\begin{document}
%

\title{Detecting Hostile Posts using Relational Graph Convolutional Network}

\titlerunning{Detecting hostile posts using Relational Graph Convolutional Networks}

%
%
\author{Sarthak \inst{1} \thanks{These two authors contributed equally}  \and
Shikhar Shukla\inst{1} \textsuperscript{*} \and
K.V. Arya\inst{2}} 
\authorrunning{Sarthak et al.}
%
\institute{Samsung Research Institute, Bangalore-560037, India \\
\email{\{sarthak.j2709,shikhar.00778\}@gmail.com} \\ \and 
ABV-Indian Institute of Information Technology \& Management, Gwalior-474015, India \\
\email{kvarya@iiitm.ac.in}}

\maketitle              
\setcounter{footnote}{0}

\begin{abstract}
This work is based on the submission to the competition Hindi Constraint conducted by AAAI@2021 for detection of hostile posts in Hindi on social media platforms. Here, a model is presented for detection and classification of hostile posts and further classify into fake, offensive, hate and defamation using Relational Graph Convolutional Networks. Unlike other existing work, our approach is focused on using semantic meaning along with contextutal information for better classification. The results from AAAI@2021 indicates that the proposed model is performing at par
with Google's XLM-RoBERTa on the given dataset. Our best submission with RGCN achieves an F1 score of 0.97 (\(7^{th}\) Rank) on coarse-grained evaluation and achieved best performance on identifying fake posts. Among all submissions to the challenge, our classification system with XLM-Roberta secured \textbf{\(2^{nd}\)} rank on fine-grained classification.


\end{abstract}

\section{Introduction}
Though Hindi is one of the most prominent languages in online communities, there has been a lack of research when it comes to identifying toxic and fake posts in Hindi. The need for such solutions has been felt more acutely in the light of the Covid pandemic. The online forums have been flooded with unproven remedies to cure and prevent the spread of the virus. False information regarding government policies such as lockdown has often led to panic and shortage of essential supplies in the communities. On the other hand, hate and offensive posts on online social forums targeting specific groups and religious beliefs have sometimes sparked violent incidents. Therefore, it is imperative to develop systems that can filter out such posts. Through the ``Hostile Post Detection in Hindi" task, the organizers have provided a manually annotated dataset\cite{bhardwaj2020hostility} which can be immensely useful in combating the above-mentioned challenges. \\
\tab The paper is organized as follows. Section 2 describes related work. Section 3 describes the background work on Relational Graph Convolutional Netowrks. Section 4 describes the methodologies of proposed models. Sections 5 presents analysis on the performance of the proposed models.
Paper is concluded is Section 6.

\section{Related Work}

Several people have presented different approaches on how to classify toxic comments on different datasets presented by Google Jigsaw during a Kaggle Competition\footnote{\url{https://www.kaggle.com/c/jigsaw-toxic-comment-classification-challenge}} and Twitter\cite{davidson2017automated}. Several authors \cite{vanaken2018challenges,gao-huang-2017-detecting} 
suggested an ensemble learning method which outperformed baselines by 7\%. However, another dataset\footnote{\url{https://www.kaggle.com/c/jigsaw-unintended-bias-in-toxicity-classification}} was released later as models were getting biased towards  certain identity words(gay, black, etc.). The top solutions used transformers to achieve state-of-the-art results on this dataset. \\
\tab However, all the models designed for them would work efficiently for classifying posts in English. Priya et al. \cite{rani2020comparative} compiled a Hindi-English code-mixed dataset for 
detecting hate speech in social media communications. They also proposed an architecture using Word2vec and FastText embeddings for classification. 
For the given dataset, organizers\cite{bhardwaj2020hostility} presented a baseline model of SVM which used multilingual embeddings as input.   \\
\tab Recently, novel architectures\cite{yao2018graph,xu2019look} using Graph Convolutional Networks \cite{kipf2017semisupervised,schlichtkrull2017modeling} have achieved state-of-the-art results for text classification on GAP \cite{webster2018gap}, Ohsumed\cite{ohsumed}, 20ng\cite{Dua:2019}, and Reuters\cite{Dua:2019} datasets. They generalize better which motivated us to use them for this task. Also, to the best of our knowledge, this is the first attempt at using Relational Graph Convolutional Networks for classifying hostile comments in Hindi.\\
\tab The recently proposed XLM-Roberta \cite{conneau2019unsupervised}, a multilingual model, outperforms all the other transformer architectures on classification tasks such as Natural Language Inference (NLI) and Named Entity Recognition (NER) in multiple languages. It was also a part of the top performing solutions in a recent Kaggle challenge \footnote{\url{https://www.kaggle.com/c/jigsaw-multilingual-toxic-comment-classification}} on identifying hostile posts in multiple languages. This motivated us to finetune the model on our dataset. 

\section{Background: Relational Graph Convolutional Networks}

\subsection{Relational Graph Convolutional Networks}

GCNs\cite{kipf2017semisupervised,BrunaGCN} are used for performing convolution over graphs. GCNs, with and without labeled edges, have been used in many text classification tasks \cite{vashishthdating,xu2019look,yao2018graph}) and achieved state-of-the-art results. Let \( \mathcal{G} = \) (\( \mathcal{V} \) , \( \mathcal{E} \)) be a directed graph where nodes
\( v_{i} \) \( \in \mathcal{V} \) and edges \( (v_{i},v_{j}) \in \mathcal{E} \) with label \(r\). 
The hidden state of each node \( {v}_{i} \) is represented as \( {h}_{i}\), where \( {h}_{i}\) is a \( {d}_{0}\)-dimensional vector. Every node \(v_{i}\) aggregates neighbours information \( {h}_{j}\) in it as described in Eq. (\ref{eq:RGCN_eq}). 

\begin{equation}
    h^{l+1}_{i} = \psi\left(\sum_{r \in R}\sum_{j \in N^r_{i}} \frac{1}{c_{i,r}} W^{(l)}_{r}h^{(l)}_{j} + B^{(l)}_{r} , \right) , \tab \forall v_{i} \in \mathcal{V}
    \label{eq:RGCN_eq}
\end{equation}

where \( \psi \) denotes an activation function, where \( c_{i,r}\) represents a normalization constant, \( N^r_{i} \) represents the degree of the node \(i\), and \( W^{(l)}_{r} \) represents the weight matrix under relation \(r\). \\

\subsection{Edge Importance(Gating Mechanism)}

In many cases, not all edges might be relevant to the model. The concept of Edge importance was introduced by Marcheggiani et al.\cite{schlichtkrull2017modeling} for helping the model in identifying the erroneous or irrelevant edges. The gating value for an edge is computed as:
\begin{equation}
    g^{(l)}_{(u,v)} = \sigma \left( h^{(l)}_{u} \cdot w^{(l)}_{(u,v)} + b^{l}_{(u,v)} \right)
\end{equation}

where \(\sigma(\cdot)\) is the sigmoid function, \( w^{k}_{(u,v)}\)  and \(b^{k}_{(u,v)}\) are gating parameters at layer \(l\). Finally, the gated R-GCN embedding or hidden state of node for our model is computed as:

\begin{equation}
    h^{l+1}_{i} = ReLU \left(\sum_{r \in R}\sum_{j \in N^r_{i}} g^{(l)}_{(i,j)} \times \left (\frac{1}{c_{i,r}} W^{(l)}_{r}h^{(l)}_{j} + B^{(l)}_{r} \right) \right) , \tab \forall v_{i} \in \mathcal{V}
    \label{eq:gated RGCN_eq}
\end{equation}

\subsection{Dependency Parsing}

\begin{figure}[H]
    \centering
    \resizebox{\textwidth}{!}{%
    \includegraphics[width=\textwidth]{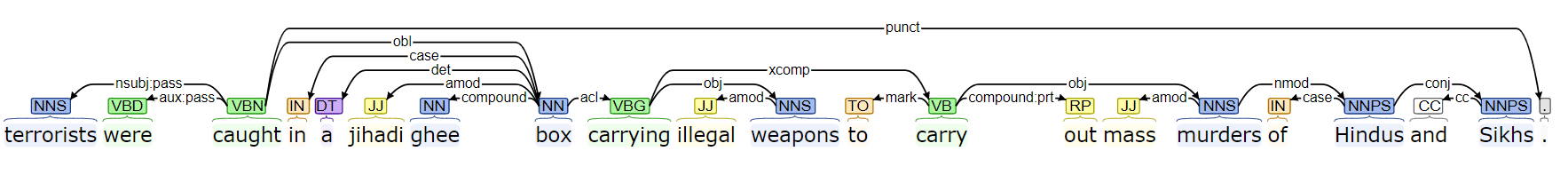}
    }
    \caption {Dependency Graph\cite{corenlp}}
    \label{fig:Dependency Graph}
\end{figure}

\tab Capturing syntactic information in a sentence helps in capturing the semantic meaning and has often been found to be useful in many tasks\cite{xu2019look,zhang2020semanticsaware} when fed into models. Fig. \ref{fig:Dependency Graph} shows a syntactic graph of a sentence.

\section{Methodology}

In this section, we present three model architectures. First subsection elaborates on proposed architecture using R-GCN and multilingual BERT. Next two subsections discuss on how to finetune pretrained architectures for this task.

\subsection{RGCN with Multilingual Bert}

\begin{figure}[H]
    \centering
    \resizebox{0.9\textwidth}{!}{%
    \includegraphics[width=\textwidth]{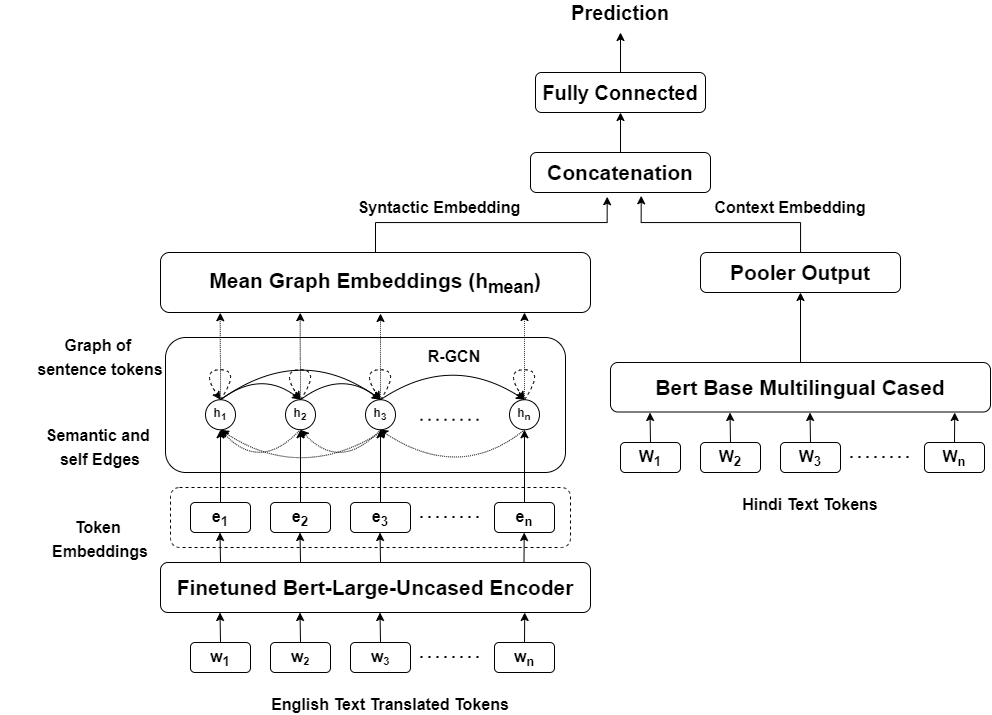}
    }
    \caption {R-GCN Architecture}
    \label{fig:R-GCN Architecture}
\end{figure}

Our deep learning-based multilabel classification architecture (as seen in Fig. \ref{fig:R-GCN Architecture}) is inspired by the approach followed by Xu et al.\cite{xu2019look} and Yao et al.\cite{zhang2020semanticsaware}. It consists of two layers in parallel for capturing the contextual\footnote{reference to person, entity or event} and semantic information of the sentence.  

\begin{itemize}
  \item \textbf{Context Embedding: } In this layer, the original Hindi text is fed into  \emph{bert-base-multilingual-cased} and pooler output is captured. It so happens, that it represents the [CLS] token in BERT which is prepended to every sentence. The embedding of this token is obtained by pooling, or making it dependent on each token in a sentence, and finetuning it using sequence classification tasks. It captures the contextual information better than semantic context. \\
  \item \textbf{Syntactic Embedding: } 
    We were not able to find dependency parsers for Hindi text. So, we translated data to English and did dependency parsing on them using spaCy\cite{spacy}. Dependency parsing is done for each sentence giving us a labeled directed graph with nodes as tokens in a sentence. As mentioned in \cite{schlichtkrull2017modeling} and \cite{yao2018graph}, we also use three types of relationships for edge labels: heads to dependents \((u,v)\), dependents to heads   
    \((u,v)^{-1}\) and self-loops \((u,u)\) where \( (u,v) \in \mathcal{E} \) and \( u,v \in \mathcal{V} \). Instead of random initialisation of hidden states of a node, we used embeddings of each token obtained from pretrained \emph{Bert-Large-Uncased}. This will help the gated R-GCN in capturing semantic task-specific embeddings. Finally, we obtain sentence semantic embedding \(h_{mean}\) by average pooling of every token embedding.  
\end{itemize}

\begin{figure}[H]
    \centering
    \resizebox{1.0\textwidth}{!}{%
    \includegraphics[width=\textwidth]{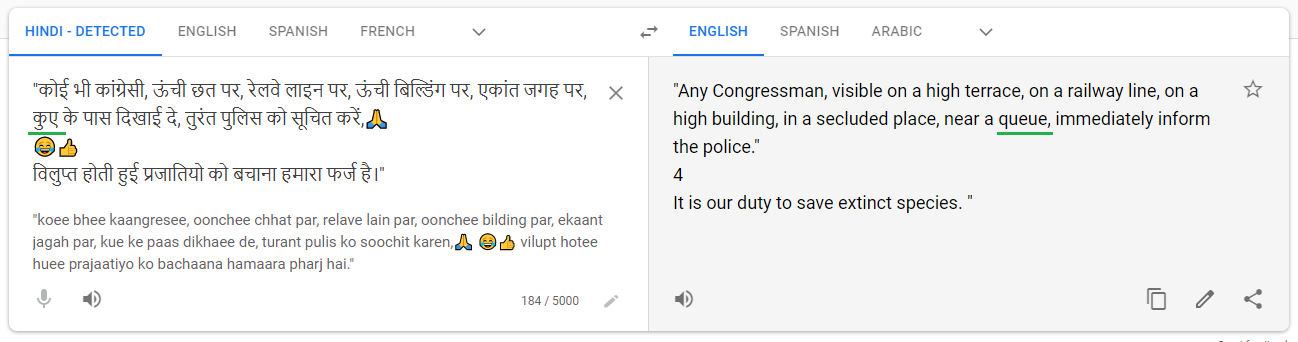}
    }
    \caption {Hindi to English Translation}
    \label{fig:Translation}
\end{figure}

The Pooler output from BERT and pooled output from Gated R-GCN is concatenated. The reason for doing this is two-fold. First, it captures both the semantic and contextual information of the text. Secondly, some contextual information might get lost in translation, which might impact semantic information as well. As evident by Fig. \ref{fig:Translation}, the semantic meaning is almost intact but a word is incorrectly translated(highlighted in green, Hindi text mentions about a ``well", but it's translated as a ``queue"). The concatenated output is then passed through a fully connected layer for final prediction. 
\\
\tab We experimented RGCN with pooler outputs from other architectures as well. The result for which can be seen in table \ref{tab:Results}.

\subsection{Training LSTM/RNN models}
\begin{itemize}

\item \textbf {Preprocessing}: The given dataset had tweets that contained unwanted tokens such as tags and shortened URLs. We cleaned these using the ``fixhtml'' rule, available as part of the Fastai\cite{Howard_2020} text processing package.\\
\item  \textbf {Tokenization}: We used sentencepiece with two tokenization schemes: \\
\begin{itemize}
\item \textbf{UnigramTokenizer} : We used a vocabulary size of 20k with unigram language model \cite{kudo2018subword}. It starts out with a seed vocabulary set, and keeps on dropping a fixed percentage of subwords\footnote{substrings in a word} to optimize the marginal likelihood till the desired vocabulary size is reached as shown in Eq. (\ref{eq:likelihood}).

\begin{equation}
    P(\mathcal{\textbf{x}}) = \prod^{M}_{i=1} p(x_{i}) \tab ; \tab \forall x_{i} \in V , \sum_{x \in V} p(x) = 1
    \label{eq:Tokenization}
\end{equation}

where , M is a subword sequence \(\mathcal{\textbf{x}} = \left(x_{1},x_{2},...x_{M} \right) \), \(V\) is the vocabulary, \(X\) is the corpus, \(S(X)\) represents a possible set of subword sequences, \(|D|\) represents number of sets in \(S\). 

\begin{equation}
    L = \sum^{|D|}_{s=1}\log(P(X^{(s)})) = \sum^{|D|}_{s=1}\log\left( \sum_{x \in S(X^{(s)})} P(\mathcal{\textbf{x}}) \right )
    \label{eq:likelihood}
\end{equation}

\item \textbf{Byte Pair Encoding: } BPE \cite{sennrich2016neural} starts out by computing the frequency of characters in the text and iteratively merges the most common pair of tokens till that point. The recently merged tokens are added to the initial list and the frequency of each token is recalculated. This is repeated till the set vocabulary limit is reached. We got better results with BPE as compared to Unigram tokenization as observed from Table \ref{tab:Results}. We used a vocabulary size of 20k.

\end{itemize}

\begin{figure}[H]
    \centering
    \resizebox{0.35\textwidth}{!}{%
    \includegraphics[width=0.3\textwidth]{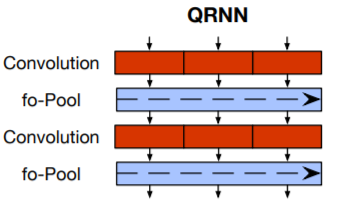}
    }
    \caption {QRNN architecture}
    \label{fig:QRNN}
\end{figure}
\item \textbf{Training}: To establish a baseline, we started by training Average-SGD Weight Dropped LSTM (AWD-LSTM) \cite{merity2017regularizing} and AWD-QuasiRNN \cite{bradbury2016quasirecurrent} models from scratch. In several tasks, these two models have performed comparable to the more recent transformer architectures. These networks use DropConnect, which sets a randomly selected subset of weights to zero. QRNN has alternate recurrent and convolutional layers, which speeds up training and testing because of parallel operations. They also outperform stacked LSTMs of similar size (see Table \ref{tab:Results}).

\end{itemize}

\subsection{Fine-tuning transformer model}

We used pretrained multilingual transformer models from Huggingface library \cite{wolf2020huggingfaces}. We finetuned XLM-Roberta Large model which has been trained on CommonCrawl data in 100 languages, with masked language modeling objective on text sampled from multiple languages. 
\begin{itemize}
\item \textbf{Dataset augmentation}: We also experimented with finetuning transformers on text from Kaggle’s Toxic Comment Classification Challenge. Then we translated the Hindi Hostile Post dataset to English using Google Translate and used these fine-tuned models for further training and classification.
Separately, we also augmented our Hindi dataset with data from HASOC (2019) challenge and finetuned XLM-Roberta on this dataset.\\
\item \textbf{Pseudo-labeling}: We used the predictions made by our trained models on the test set as soft labels and retrained the model after upsampling this data. Data augmentation and pseudo-labeling didn't provide any performance boost to our models. 

\end{itemize}

\section{Results and Analysis}

\subsection{Dataset}

\begin{figure}[H]
\begin{subfigure}{.3\textwidth}
  \centering
  \includegraphics[width=0.8\linewidth]{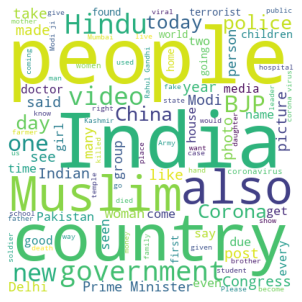}  
  \caption{Fake}
  \label{fig:fake}
\end{subfigure}
\begin{subfigure}{.3\textwidth}
  \centering
  \includegraphics[width=0.8\linewidth]{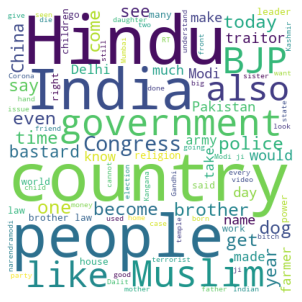}  
  \caption{Hate}
  \label{fig:hate}
\end{subfigure}
\begin{subfigure}{.3\textwidth}
  \centering
  \includegraphics[width=0.8\linewidth]{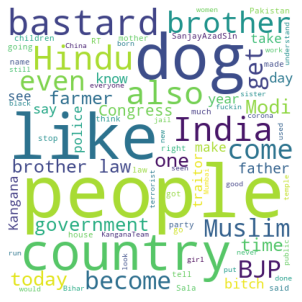}  
  \caption{Offensive}
  \label{fig:offensive}
\end{subfigure}
\label{fig:fig}
\hspace{70px}
\begin{subfigure}{.3\textwidth}
  \centering
  \includegraphics[width=0.8\linewidth]{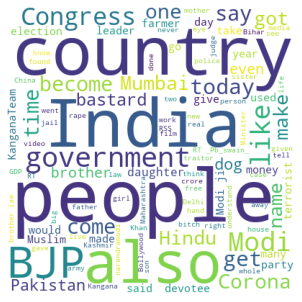}  
  \caption{Defamation}
  \label{fig:Defamation}
\end{subfigure}
\begin{subfigure}{.3\textwidth}
  \centering
  \includegraphics[width=0.8\linewidth]{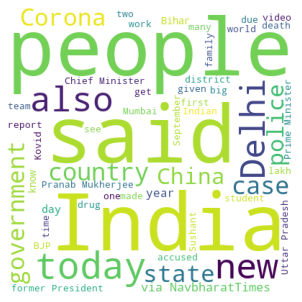}  
  \caption{Non-hostile}
  \label{fig:non}
\end{subfigure}
\caption{Word Clouds}
\label{fig:Word Cloud}
\end{figure}

The training and validation datasets\cite{bhardwaj2020hostility} were provided as part of the competition, which were split into hostile and non-hostile classes. The hostile posts were further divided into fake, hate, offensive and defamation categories. The training and validation sets had 5728 and 811 samples respectively. The comments had an average length of 30 tokens. We have presented word clouds for translated data of each class in Fig. \ref{fig:Word Cloud}.

\subsection{Evaluation}

The metric for evaluation in this task was F1 score. Since the data suffers from class imbalance and its important to maintain a balance between precision and recall, weighted F1 scores become a better metric for evaluation. 
We evaluate our models\footnote{\url{https://github.com/shikhar00778/constraint21}} on the test set and the F1 scores are mentioned in Table \ref{tab:Results}. 

\begin{table}[H]

\bgroup
\setlength{\tabcolsep}{4pt} 
\def\arraystretch{1.2}%
\centering
\caption{Coarse and fine grained F1 scores}
\label{tab:Results}
\resizebox{1.0\textwidth}{!}{%
 \begin{tabular}{||c c c c c c c||} 
 \hline
 \textbf{Model} & \textbf{Coarse} & \multirow{2}{*}{\textbf{Fake}} & \multirow{2}{*}{\textbf{Hate}} & \multirow{2}{*}{\textbf{Defamation}} & \multirow{2}{*}{\textbf{Offensive}} & \textbf{Weighted Fine}\\ 
 \textbf{Details} & \textbf{Grained}&  & & &  & \textbf{Grained}\\ 
 \hline\hline 
 XLM-Roberta\tablefootnote{bestfit\_ai 3$^{rd}$ submission} & 96.610 & 82.082 & 55.870 & \textbf{44.982} & 59.447 & \textbf{63.922}\\[0.5 ex] 
 RGCN + BERT\tablefootnote{bestfit\_ai 2$^{nd}$ submission} & \textbf{96.614} & \textbf{82.440} & \textbf{58.555} & 31.538 & 58.951 & 62.214\\[0.5 ex]
 RGCN + XLM-RoBERTa & 96.305 & 78.865 & 54.0 & 40.234 & 54.242 & 60.343\\[0.5 ex]
 AWD-LSTM & 96.4 & 74.2 & 52.7 & 38.4 & 59.7 & 53.7\\[0.5 ex]
 AWD-QRNN & 96.1 & 60.6 & 53.7 & 39.5 & \textbf{60.6} & 53.3\\ [0.5 ex]
 \hline
 \end{tabular}
 }
 
\egroup
\end{table}

\begin{figure}
\begin{subfigure}{1.0\textwidth}
  \centering
  \includegraphics[width=1.0\linewidth]{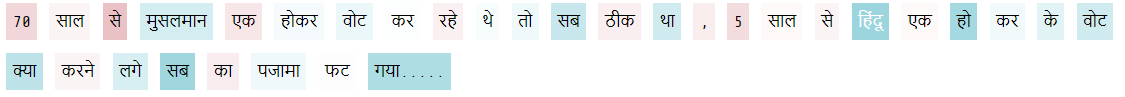}  
  \subcaption{Hate}
  \label{fig:hate_lit}
\end{subfigure}
\begin{subfigure}{1.0\textwidth}
  \centering
  \includegraphics[width=1.0\linewidth]{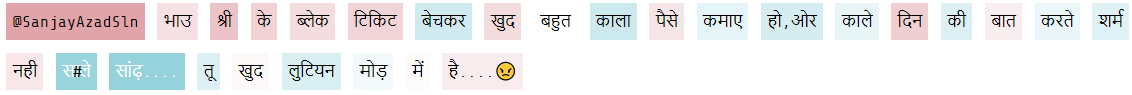}  
  \subcaption{Offensive}
  \label{fig:offensive_lit}
\end{subfigure}
\begin{subfigure}{1.0\textwidth}
  \centering
  \includegraphics[width=1.0\linewidth]{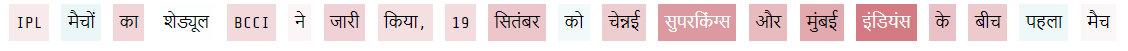}  
  \subcaption{Non-hostile}
  \label{fig:non_hostile_lit}
\end{subfigure}
\label{fig:}
\caption{Salience maps for posts with XLM-Roberta (Red tokens are less important, blue tokens are more significant to the model in the final prediction)}
\label{fig:Saliency maps}
\end{figure}

\begin{figure}[H]
        \centering
        
        \captionsetup{justification=centering}
        \begin{subfigure}[b]{0.475\textwidth}
            \centering
            \resizebox{0.9\textwidth}{!}{%
            \includegraphics[width=\textwidth]{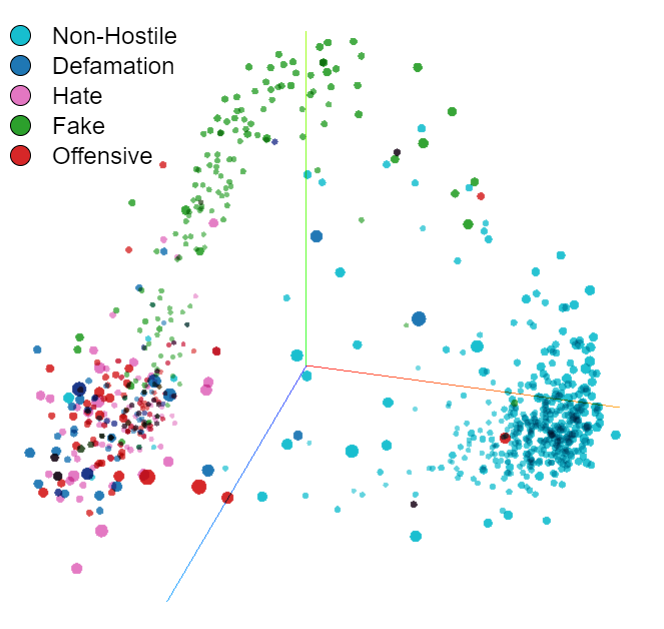}
            }
            \caption[]%
            {{\small Last layer embeddings from trained XLM-Roberta}}    
            \label{fig:wxlm}
        \end{subfigure}
        \hfill
        \begin{subfigure}[b]{0.475\textwidth}  
            \centering 
            \resizebox{0.9\textwidth}{!}{%
            \includegraphics[width=\textwidth]{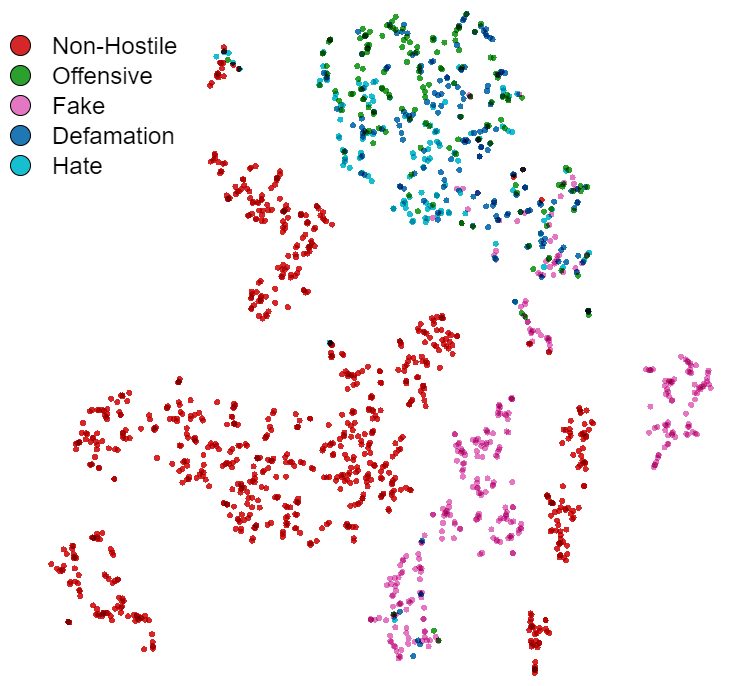}
            }
            \caption[]%
            {{\small Pooler Output from multilingual BERT}}    
            \label{fig:pooler output mbert}
        \end{subfigure}
        \vskip\baselineskip
        \begin{subfigure}[b]{0.475\textwidth}   
            \centering 
            \resizebox{0.9\textwidth}{!}{%
            \includegraphics[width=\textwidth]{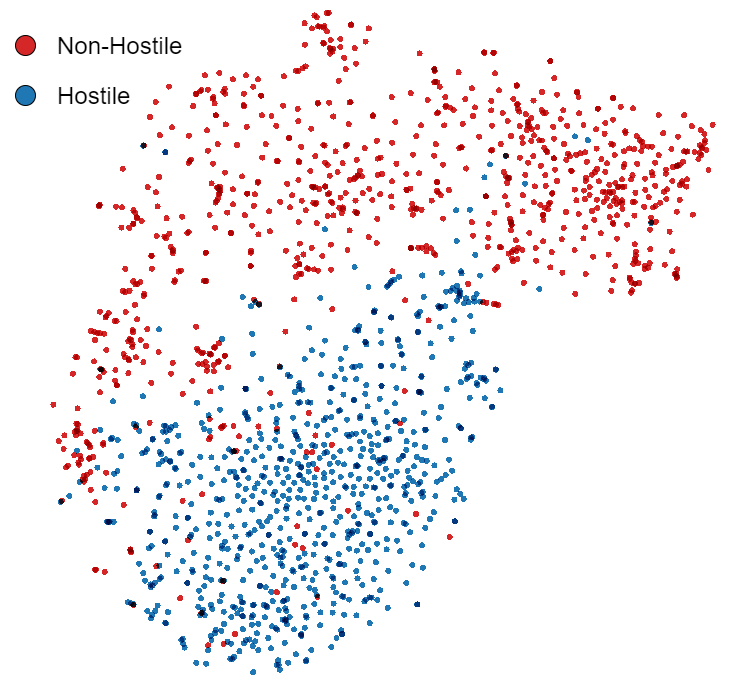}
            }
            \caption[]%
            {{\small Mean Sentence Embeddings from trained RGCN layer}}    
            \label{fig:rgcn embed}
        \end{subfigure}
        \hfill
        \begin{subfigure}[b]{0.475\textwidth}   
            \centering 
            \resizebox{0.9\textwidth}{!}{%
            \includegraphics[width=\textwidth]{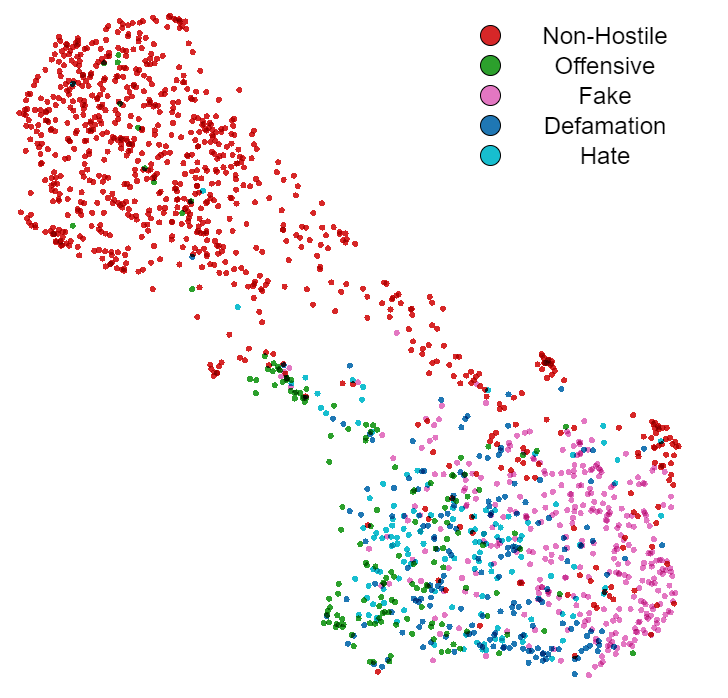}
            }
            \caption[]%
            {{\small Concatenated embeddings of RGCN and BERT}}    
            \label{fig:concat embed}
        \end{subfigure}
        \caption[]
        {\small Embedding Visualisations of XLM-Roberta and RGCN+BERT} 
        \label{fig:embed visualisation}
\end{figure}

We extracted last layer embedding from the trained XLM-Roberta model and visualized it on validation set, after using PCA as shown in Fig. \ref{fig:wxlm}. The plot illustrates why the model performs well in classifying non-hostile and fake posts. Also, the embeddings for hate, offensive and defamation posts are clustered together, thus resulting in a poor performance on those classes. In Fig. \ref{fig:Saliency maps}, we visualize the salience maps of a few samples from the validation set, using Google's Language Interpretability Tool\cite{tenney2020language} with XLM-Roberta. It depicts each token's contribution to the final prediction made by the model. The tokens in red are less important, while the model focuses more on the blue tokens. The plot \ref{fig:hate_lit} demonstrates how the model has learnt to attach more weight to words such as ``Hindu" and ``Muslim". These words tend to appear often in hostile posts in the given dataset.

\begin{figure}[H]
    \centering
    \resizebox{0.9\textwidth}{!}{%
    \includegraphics[width=0.3\textwidth]{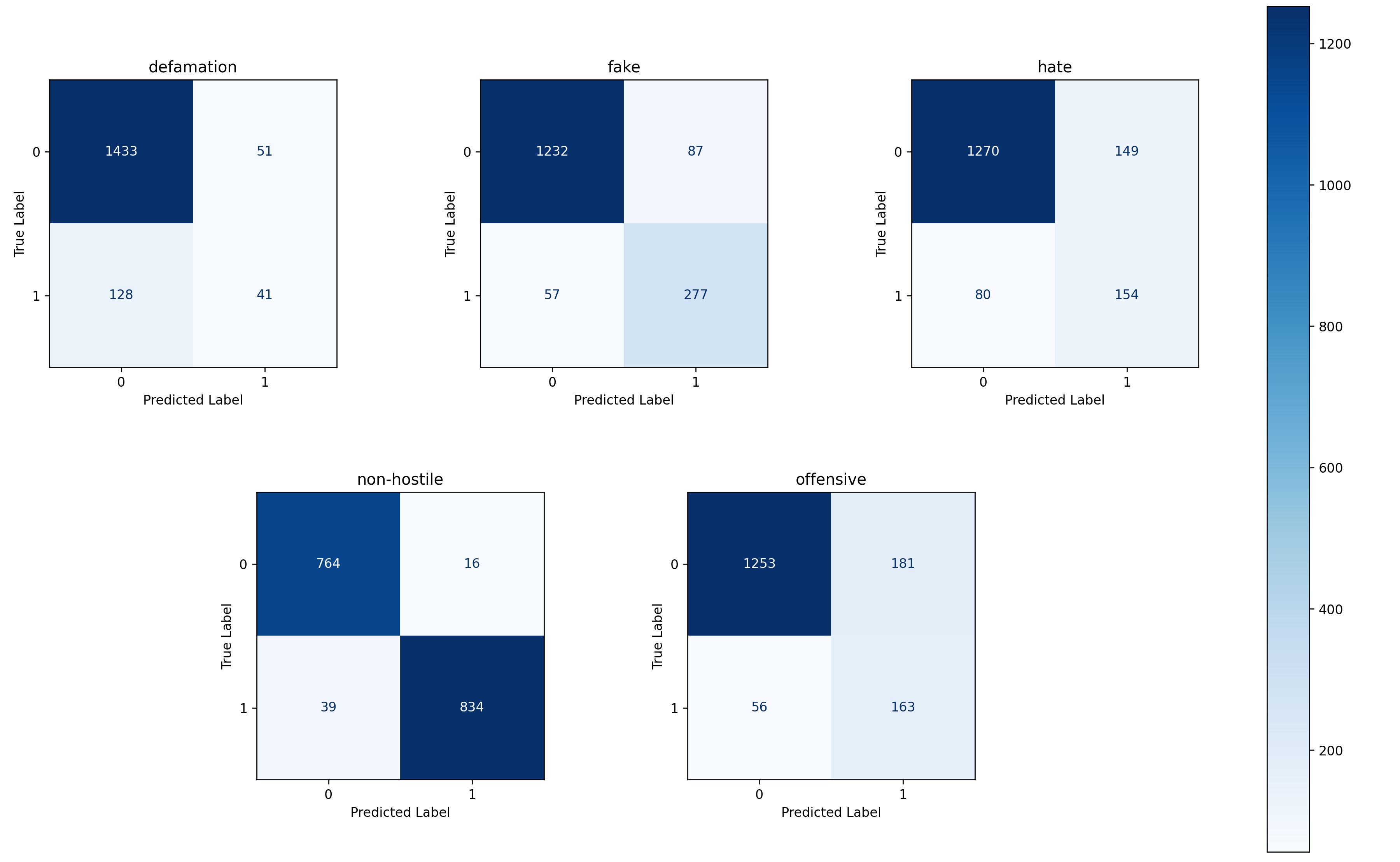}
    }
    \caption {Confusion Matrix for classification using R-GCN}
    \label{fig:confusion RGCN}
\end{figure}

Sentence embeddings from RGCN layer, multilingual BERT and concatenated embeddings from the trained RGCN-BERT model were also extracted and visualized on test set using T-SNE as shown in Fig. \ref{fig:embed visualisation}. The plot \ref{fig:rgcn embed} illustrates why the model performs well in achieving high coarse grained score. 
Fig. \ref{fig:pooler output mbert} illustrates how multilingual BERT can clearly classify fake news from other classes. Fig. \ref{fig:concat embed} illustrates the final sentence embeddings obtained after concatenating the RGCN and multilingual BERT embeddings. Fig \ref{fig:confusion RGCN} presents the confusion matrix of trained RGCN+BERT model on test dataset for all the categories. Overall, model can clearly identify non-hostile and fake posts due to the advantages that RGCN and BERT embeddings carry, and why it has outperformed all other submissions for classifying fake posts.

\section{Conclusion}
In this paper, a model is presented for detecting hostile posts and further classifying them into fake, hate, offensive and defamation. By combining semantic information along with the contextual information leading to improved performance was observed. The proposed model achieved (\(5^{th}\) Rank) on fine-grained evaluation in Hindi Constraint organized by AAAI@2021. RGCN with Bert performed better than all the other teams' submissions to the challenge on classifying fake posts, achieving an F1 score of 82.4. Limitation of the proposed model is that it is not able to classify defamation posts with the best F1 score of 44.9 across all our team's submissions.\\
\textbf{Future Scope:}  The work can further be explored by doing syntactic analysis directly on Hindi text rather than translated text. Furthermore, finetuning multilingual T5 for this task can also yield better results. Also, ensembling the results of XLM-Roberta and RGCN+Bert architecture might give better results since the former performed well on defamation while the latter achieved state-of-the-art results for fake news detection. 
%
%
%
%

\bibliographystyle{splncs04}
\bibliography{references}

\end{document}